\documentclass[journal]{IEEEtran}
\ifCLASSINFOpdf
\else
\fi

\usepackage{color}
\usepackage[table,usenames,dvipsnames]{xcolor}
\usepackage{graphicx}

\usepackage{amsmath}
\usepackage{mathtools}
\usepackage{placeins} 

\usepackage{algpseudocode}
\usepackage{algorithm}
\usepackage{multirow}

\usepackage{cite}

\definecolor{linkcolor}{RGB}{0,0,128}

\usepackage[pdfpagelabels,
            plainpages=false,
            hypertexnames=true,
            colorlinks=true,
            linkcolor=linkcolor,
            anchorcolor=linkcolor,
            citecolor=linkcolor,
            filecolor=linkcolor,
            menucolor=linkcolor,
            runcolor=linkcolor,
            urlcolor=linkcolor,
            pdfborder={0 0 0} ,
            hidelinks]{hyperref}

\usepackage[
            color=green!10,
            colorinlistoftodos,
            linecolor=lightgray,
            bordercolor=green!10,
            textsize=tiny
            ]{todonotes} 

\usepackage{dblfloatfix}

\usepackage{xargs} 
\newcommandx{\dwsay}[2][1=]{\todo[
            color=SkyBlue!50,
            bordercolor=SkyBlue!50,
            linecolor=lightgray,#1]{\emph{DW}: #2}}
\newcommandx{\ipsay}[2][1=]{\todo[
            color=SeaGreen!50,
            bordercolor=SeaGreen!50,
            linecolor=lightgray,#1]{\emph{IP}: #2}}

\begin{document}
%
\title{Marginal likelihood based model comparison in Fuzzy Bayesian Learning}

%
%
%


\author{Indranil~Pan
        and~Dirk Bester

\thanks{I. Pan was with Imperial College London, UK, and now works at Sciemus Ltd, London UK. e-mail: ipan@sciemus.com}
\thanks{D. Bester was with University of Oxford, UK, and now works at Sciemus Ltd, London UK. e-mail: dbester@sciemus.com}}
\maketitle

\begin{abstract}
In a recent paper \cite{pan2016fuzzy} we introduced the Fuzzy Bayesian Learning (FBL) paradigm where expert opinions can be encoded in the form of fuzzy rule bases and the hyper-parameters of the fuzzy sets can be learned from data using a Bayesian approach. The present paper extends this work for selecting the most appropriate rule base among a set of competing alternatives, which best explains the data, by calculating the model evidence or marginal likelihood. We explain why this is an attractive alternative over simply minimizing a mean squared error metric of prediction and show the validity of the proposition using synthetic examples and a real world case study in the financial services sector.

\end{abstract}


%

\section{Introduction}
%
%
%
%

\IEEEPARstart{T}{raditional} rule based fuzzy systems encode expert opinion in the form of IF-THEN rules and optimise some performance metric (typically mean squared error in predicting a data-set) to obtain the hyper-parameters of the model (like the numeric values defining the shape of the membership functions etc.) \cite{juang2014rule,cara2013multiobjective,gil2015gain}.
The rule base is one of the core elements driving the model and slight change in the rule base would significantly affect the performance of the fuzzy inference system. Many automated methods have been proposed where the rule base structure and parameters can be automatically determined \cite{zhao2013hybrid,othman2014efis,lin2015interval}. However interpretability of such models is an issue and various methods have been proposed to alleviate the issue \cite{zhou2008low}. In the present paper however, we are interested in the actual metric for comparison between different models having different rule bases derived from expert opinion. The comparison metric can nevertheless be embedded within an automated framework to evolve the best model if so required.   

The most straight forward way of comparing the fuzzy rule bases is to optimise the model parameters based on the prediction error (e.g. mean squared error, mean absolute error etc.) on a data-set and to select the rule base with the lowest error as the better model, for the given data-set. Improvisations on this method would be to divide the data into a training and testing set and use the average prediction from k-fold cross validation on the data-set. However, as demonstrated in \cite{piironen2015comparison}, the optimisation of an utility estimate like cross validation, with scarce data (which is the case for us), is likely to find overfitted models due to high variance in the utility estimates. The authors also show that Bayesian model averaging over all the candidate models takes the model uncertainty into account and is less prone to overfitting than cross-validation. A more generic overview of the different methods of Bayesian model comparison is presented in \cite{vehtari2012survey}.   

Another popular method is to use information criteria (IC) like Akaike's, Deviance, Bayesian information criteria etc. for model comparison, which penalises both the prediction misfit and the complexity of the model. Now, the complexity of the model is generally penalised based on the number of effective parameters ($N_{\text{eff}}$) \cite{lunn2012bugs}; the rationale being that models with more free parameters would have better flexibility at fitting the data and hence more complexity. But this is not necessarily true for all class of models. For example, the fuzzy models considered in this paper have the same number of parameters to be estimated (as the number of membership functions are the same), but differ in the number and definition of rule bases. The rule bases essentially act like a complicated non-linear kernel to transform the input covariates into the output. This would be difficult to capture in $N_{\text{eff}}$ and would need customised application specific methods.  

All the ICs are essentially approximations of the marginal likelihood or the Bayesian model evidence since the corresponding integral is very hard to estimate \cite{murphy2012machine}. A rigorous comparison of different ways to evaluate Bayesian model evidence is shown in \cite{schoniger2014model} and it is demonstrated that Bayesian evidence approximations from ICs are often heavily biased and accuracy of the model ranking is significantly affected by the choice of the approximation method. The authors also suggest using bias free numerical methods (e.g. nested sampling \cite{skilling2004nested}), over ICs if it is computationally feasible. The marginal likelihood computation integrates out model parameters rather than maximising them, which automatically prevents overfitting \cite{murphy2012machine}. It also maintains a trade-off between model complexity and fit to the dataset. Also the marginal likelihood can be factored into a product of predictive likelihoods which reflect the model's ability to make predictions on new data. 
These reasons make the marginal likelihood an attractive metric for model selection.   

In the next sections we briefly summarise the mathematical foundations of Bayesian model selection, the basic nested sampling algorithm with its improvements, integrate them with the Fuzzy Bayesian Learning framework and show demonstrative examples on synthetic and real world data-sets.  

\vfill

\section{Bayesian evidence for model selection in FBL and nested sampling}
\subsection{Bayesian evidence and model selection using nested sampling}

Bayes theorem allows for the calculation of the posterior probability of a model (or hypothesis), from a set of plausible ones, given the data and can be expressed as

\begin{equation}
\label{eq:myeqdef1}
p\left( H_i \mid \boldsymbol{D} \right)
=
\frac{
       p\left( \boldsymbol{D} \mid H_i \right)
       p\left( H_i \right)
     }{
       p\left( \boldsymbol{D} \right)
      }
\end{equation}
where $p\left( H_i \right)$ is the prior probability of the $i^{th}$ hypothesis ($H$) and $p\left( \boldsymbol{D} \right)$ is the probability of the data.
The posterior probability of the parameters ($\theta$) of $H_i$ given $\boldsymbol{D}$ is
\begin{equation}
\label{eq:myeqdef2}
p\left( \theta \mid \boldsymbol{D}, H_i \right)
=
\frac{
       p\left( \boldsymbol{D} \mid \theta, H_i \right)
       p\left( \theta \mid H_i \right)
     }{
       p\left( \boldsymbol{D} \mid H_i \right)
      }
\end{equation}
where $ p\left( \boldsymbol{D} \mid \theta, H_i \right)$ is the likelihood of the data given the hypothesis and its parameters, $p\left( \theta \mid H_i \right)$ is the prior probability of the parameters given the hypothesis. These two terms in the numerator are relevant for the parameter estimation problem. For the case of model selection, the important term is the denominator $ p\left( \boldsymbol{D} \mid H_i \right)$, also known as the evidence or the marginal likelihood of the model $H_i$ \cite{mackay2003information}. The evidence ($\mathcal{Z}_i$), for the $i^{th}$ model is also the normalizing constant and can be written as 
\begin{equation}
\label{eq:evidence}
\mathcal{Z}_i=p\left( \boldsymbol{D} \mid H_i \right)
= \int{ p\left( \boldsymbol{D} \mid \theta, H_i \right) p\left( \theta \mid H_i \right) d \theta}
\end{equation}

The evidence automatically implements Occam's razor \cite{smith1980bayes,jefferys1992ockham,mackay1992bayesian}, which implies that a simpler model with less number of parameters would have a higher value of evidence than a more complex model, unless the latter can explain the data much better. The model selection problem between two competing models (hypothesis) $H_1$, $H_2$ can be expressed as the ratio of how probable the respective models are in light of the given data-set $\boldsymbol{D}$ (i.e. the ratio of their posterior distributions), given by the following equation

\begin{equation}
\label{eq:modelcomp}
\frac{
       p\left( H_2 \mid  \boldsymbol{D}\right)
     }{
       p\left( H_1 \mid \boldsymbol{D}\right)
      }
=
\frac{
       p\left( \boldsymbol{D} \mid H_2 \right)
       p\left( H_2 \right)
     }{
       p\left( \boldsymbol{D} \mid H_1 \right)
       p\left( H_1 \right)
      }
=
\frac{
       \mathcal{Z}_2
       p\left( H_2 \right)
     }{
      \mathcal{Z}_1
       p\left( H_1 \right)
      }
\end{equation}

which essentially reduces to the ratio of the model evidences, if the prior probabilities of both the models are the same (i.e. $p\left( H_2 \right)=p\left( H_1 \right)$). 

The evidence in Eq. (\ref{eq:evidence}) is a multi-dimensional integral and has been traditionally difficult to handle with numerical approximation techniques. The canonical method is to apply thermodynamic integration which employs a sophisticated version of MCMC sampling. However it requires large number of function evaluations (at least an order of magnitude more than parameter estimation) \cite{feroz2008multimodal} and cannot handle phase changes properly \cite{skilling2004nested}. A better way to evaluate the evidence is through the use of nested sampling \cite{skilling2004nested} which overcomes the afore-mentioned problems. 
It employs two key concepts -- the first is that the multi-dimensional evidence integral in Eqn. (\ref{eq:evidence}) can be transformed into a one dimensional integral. Following the lines of \cite{skilling2004nested}, the total prior mass is denoted by $X$ and each unit in the prior mass as $dX$ where

\begin{equation}
\label{eq:dX}
p\left( \theta \mid H \right)d\theta=\pi\left(\theta \right)d\theta=dX
\end{equation}

The likelihood can then be expressed as
\begin{equation}
\label{eq:lik}
p\left( \boldsymbol{D} \mid  \theta, H \right)=L\left(\theta \right)=L\left(X \right)
\end{equation}
which transforms the evidence into a one dimensional integral 

\begin{equation}
\label{eq:1d_evidence}
\mathcal{Z}=\int {L\left(\theta \right) \pi\left(\theta \right)}d\theta=\int_0^1 {L\left(X \right)}dX
\end{equation}
summing over the prior mass

\begin{equation}
\label{eq:sum_prior}
X\left(\lambda \right)=\int_{L\left(\theta \right) \geq \lambda} \pi \left(\theta \right) d\theta, 
X\left(0\right)=1,X\left(\infty \right)=0 
\end{equation}
where $\lambda$ is the likelihood contour. The other key concept is that if $N$ particles $\theta_1, \dots ,\theta_N$ explore the prior above the likelihood contour $\lambda$

\begin{equation}
\label{eq:lik_contour}
\begin{array}{cc}
\theta_n \sim p\left(\theta \mid \lambda \right)= \frac{\Theta [L(\theta)-\lambda]}{X(\lambda)}\pi(\theta) \\
\text{where }
\Theta(x)=\left\{
                      \begin{array}{ll}
                      0; x <0 \\
                      1; x \geq 0 \\
                      \end{array}
                      \right.
\end{array}
\end{equation}
the corresponding prior masses $X_n \equiv X(L_n)$ enclosed by the contours $L_n \equiv L(\theta_n)$ follow an uniform distribution. $p\left(\theta \mid \lambda \right)$ in Eqn. (\ref{eq:lik_contour}) is the prior with a likelihood constraint $\lambda$. The prior mass can be ordered as $X\left(\lambda \right) < X\left(\lambda^\prime \right)$ if $\lambda > \lambda^\prime$ and the prior mass enclosed by the worst state/sample can be predicted.

In each iteration, the nested sampling algorithm moves from the current likelihood contour ($\lambda$) to the next $\lambda^\prime$, where $\lambda^\prime > \lambda$ and the relative entropy ($H$) in Eqn. (\ref{eq:rel_entropy}) can be used to measure the overlap between successive truncated priors  $p\left(\theta \mid \lambda \right)$ and $p\left(\theta \mid \lambda^\prime \right)$ \cite{skilling2004nested}. 

\begin{equation}
\label{eq:rel_entropy}
H(\lambda \rightarrow \lambda^\prime)=\int{p(\theta \mid \lambda^\prime)ln\left[\frac{p(\theta \mid \lambda^\prime)}{p(\theta \mid \lambda)}\right]}d\theta=ln\left[\frac{X(\lambda)}{X(\lambda^\prime)}\right]
\end{equation}

The average relative entropy is constant and is dependent on the number of particles $N$, fed in by the user. 
\begin{equation}
\label{eq:avg_rel_entropy}
\langle H(\lambda \rightarrow \lambda^\prime) \rangle=-\langle ln t \rangle_{t \sim Beta(N,1)}=1/N
\end{equation}

The basic nested sampling algorithm pseudo-code is shown in Algorithm 1.

 
\begin{algorithm}
\caption{Basic Nested Sampling Algorithm}
\begin{algorithmic}[1]
\renewcommand{\algorithmicrequire}{\textbf{Input:}}
\renewcommand{\algorithmicensure}{\textbf{Output:}}
 \Require $N$ (particles / live points), $tol$ (tolerance for convergence)
 \Ensure  $\mathcal{Z}$ (model evidence), posterior samples of $\theta$
\State $\lambda_0 \leftarrow 0, X_0 \leftarrow 1, S=\{ \theta_1, \dotsb ,\theta_N \}$,
$\mathcal{Z} \leftarrow 0$

\\ where $\theta_n \sim \pi(\theta) , i \leftarrow 0 $ 
\Comment Initialisation
\Repeat
\State $\theta_i^* \leftarrow \text{argmin}\{L(\theta_n) \mid n=1, \dotsb, N\}$
\Comment Find $\theta$ with smallest likelihood
\State $\lambda_i \leftarrow L(\theta_i^*)$
\Comment Assign new likelihood contour
\State $X_i \leftarrow tX_{i-1}$ \\ where $t \sim Nt^{N-1}$
\Comment Calculate prior mass
\State $\theta^\prime \sim p(\theta \mid \lambda_i)$
\Comment Sample from truncated prior
\State $S \leftarrow S \setminus \{\theta_i^*\} \cup \{\theta^\prime\}$
\Comment Replace with sampled value
\State $\mathcal{Z} \leftarrow \mathcal{Z} + \lambda_i(X_{i-1}-X_i) $
\Comment Sum evidence
\State $i \leftarrow i+1$
\Until{$tol$ reached}

\end{algorithmic}
\end{algorithm}

The key bottleneck in implementing Algorithm 1, is sampling from the truncated prior $p(\theta \mid \lambda)$. Various versions of nested sampling have evolved to solve this issue. The single ellipsoidal nested sampling \cite{mukherjee2006nested} approximates the iso-likelihood contour by a hyper ellipsoid and works well for uni-modal posteriors, but not multi-modal ones. Multi-modal nested sampling \cite{feroz2008multimodal} improves on this using multiple ellipsoids and a clustering algorithm to identify isolated modes in the posterior distribution. This works on low to moderate dimensional parameter spaces and other techniques relying on MCMC, like polychord \cite{handley2015polychord} and diffusive nested sampling \cite{brewer2011diffusive} have been proposed for high dimensional parameter spaces. 

\subsection{Application to Fuzzy Bayesian Learning}

Following on from the lines of \cite{pan2016fuzzy} the inference problem is that of estimating $\theta$ for the underlying non-linear model function $g(\boldsymbol{x};\theta)$ given the data $\boldsymbol{D}$ which comprises of the set of input vectors 
$\boldsymbol{X}_{N}=\boldsymbol{x}_{i}  \forall i \in \left\{ 1,2,...,N \right\}$
and the set of outputs 
$\boldsymbol{Y}_{N}=\boldsymbol{y}_{i}  \forall i \in \left\{ 1,2,...,N \right\}$. As introduced in \cite{pan2016fuzzy}, the function $g(\boldsymbol{x};\theta)$ is a fuzzy inference system given by the rule base Eq.  (\ref{eq:fuzzyrulebase})
\begin{equation}
\label{eq:fuzzyrulebase}
\beta_{k}R_{k}: \text{if } A^{k}_{1} \oplus A^{k}_{2} \oplus\dotso \oplus A^{k}_{T_{k}} \text{ then }  C_{k}
\end{equation}
where $\beta_{k}$ is a dichotomous variable indicating the inclusion of the $k^{th}$ rule in the fuzzy inference system, $A^{k}_{i} \;\forall i \in \left\{ 1,2,...,T_{k} \right\}$ is a referential value of the $i^{th}$ antecedent attribute in the $k^{th}$ rule, $T_{k}$ is the number of antecedent attributes used in the $k^{th}$ rule, $C_{k}$ is the consequent in the $k^{th}$ rule, $\oplus\in\left\{ \vee,\wedge \right\}$ represents the set of connectives (OR, AND operations) in the rules. A referential value is a member of a referential set with the universe of discourse $U_i$, which are meaningful representations for specifying an attribute using subjective linguistic terms which are easily understood by an expert. Each of these referential values are characterised by membership functions $f_{i}\left( A_{i},U_i, \phi_{i}\right) or f_{i}\left( C_{i}, U_i, \phi_{i}\right)$ and for the inference of the function $g(\boldsymbol{x};\theta)$ in Eq. (\ref{eq:myeqdef1}), the set of parameters to be estimated is 
$\theta$, which contains all $\phi_{i}$ and $\beta_{k}$

We assume a Gaussian likelihood function given by

\begin{equation}
\label{eq:normal:assumption}
\boldsymbol{Y}_N \sim  \mathcal{MVN} \left( \boldsymbol{\mu} , \Sigma \right)
\text{,}
\end{equation}
where
$\mathcal{MVN} (\boldsymbol{\mu} , \Sigma)$
is the Multivariate-Normal distribution with mean vector $\boldsymbol{\mu}$ and covariance matrix $\Sigma$.
The mean vector for our model comes from $g\left(\boldsymbol{X}_N;\theta \right)$ and the fuzzy inference system from Eq. (\ref{eq:fuzzyrulebase}).
We use $I_N$, the $N \times N$ identity matrix, along with a parameter $\sigma$ to construct the covariance matrix.
This leads to the likelihood
\begin{equation}
\label{eq:likelihood}
\begin{aligned}
&p\left( \boldsymbol{y}_{N}\mid g(\boldsymbol{x};\theta), \boldsymbol{X}_{N} \right)\\
&=
\frac{1}{\sqrt{(2\pi)^{N}\sigma^{2N} }} \times \\
&\phantom{=}
\exp\left(-\frac{1}{2\sigma^{2N} }
         \big(\boldsymbol{y}_{N}-g(\boldsymbol{X}_N;\theta)\big)^\mathrm{T}
         \big(\boldsymbol{y}_{N}-g(\boldsymbol{X}_N;\theta)\big)
\right)
\text{.}
\end{aligned}
\end{equation}
where $\boldsymbol{y}_N$ is a realisation of the random variable $\boldsymbol{Y}_N$.

Now for comparing different fuzzy models (i.e. different $H_i$), the components of the fuzzy inference system (e.g. rule bases, membership functions etc.) would be different, giving rise to a different $g_i(\cdot)$ for each case, along with the corresponding set of hyper-parameters $\theta$. For all the demonstrative simulations $\beta_k=1, \forall k$.  

\section{Demonstrative examples}
\subsection{Examples with a synthetic dataset}

Consider a synthetic example similar to \cite{pan2016fuzzy} where the downtime of an engineering operation needs to be predicted from two covariates - location risk and maintenance level.  For the  referential sets $loc\_risk,  maintenance,  downtime$ the referential values are $ \left\{ LO, MED, HI \right\}$ , $\left\{ POOR, AVG, GOOD\right\}$, $\left\{ LO, MED, HI\right\}$ respectively. Also consider another spurious covariate $dummy\_covar$ with referential values $\left\{ POOR, AVG, GOOD\right\}$ which is also to be tested for inclusion in model.

The corresponding membership functions are considered to be triangular and can be mathematically expressed as 

\begin{equation}
\label{eq:triang_mem}
f_i\left( u , \phi_{i} \right)= \left\{
\begin{array}{cl}
\frac{\left ( u-\phi_{i_1} \right )}{\left ( \phi_{i_2}-\phi_{i_1} \right )} & \text{if } \phi_{i_1}\leq u\leq \phi_{i_2} 
\\ 
\frac{\left ( u-\phi_{i_3} \right )}{\left ( \phi_{i_2}-\phi_{i_3} \right )} & \text{if } \phi_{i_2}\leq u\leq \phi_{i_3} 
\\
0 & \text{otherwise}
\end{array}
\right.
\end{equation}
where, $ u \in U_i $ is any value within the universe of discourse and $ \phi_{i_j} \in \phi_{i} \forall j $ are the parameters of the membership functions to be estimated by the nested sampling algorithm.

Consider the following different rules which would be combined to make up all the different models for evidence calculation.

\begin{equation}
\label{eq:rule123}
\begin{split}
\textcolor{Blue}{R_{1}}:if &\:loc\_risk==HI \; \vee \; maintenance==POOR \\
                &then \; downtime==HI \\
\textcolor{Blue}{R_{2}}:if &\:loc\_risk== MED \; \vee \; maintenance==AVG \\
                &then \; downtime==MED\\
\textcolor{Blue}{R_{3}}:if &\:loc\_risk==LOW \; \wedge \; maintenance==GOOD \\
                &then \; downtime==LOW\\ 
\textcolor{BrickRed}{R_{4}}:if &\:loc\_risk==LO \;  \\
                &then \; downtime==HI \\
\textcolor{BrickRed}{R_{5}}:if &\:maintenance==POOR \\
                &then \; downtime==LOW\\
\textcolor{ForestGreen}{R_{6}}:if &\:dummy\_covar==POOR  \\ 
                &then \; downtime==LOW\\ 
\end{split}
\end{equation}

The different models or hypotheses ($H$) considered for this example consist of the following sets of individual rules

\begin{equation}
\label{eq:models}
\begin{split}
H_{true}: \{\textcolor{Blue}{R_{1}},\textcolor{Blue}{R_{2}},\textcolor{Blue}{R_{3}}\} \\
H_{1}: \{\textcolor{Blue}{R_{1}},\textcolor{Blue}{R_{2}},\textcolor{Blue}{R_{3}}\, \textcolor{BrickRed}{R_{4}},\textcolor{BrickRed}{R_{5}}\} \\
H_{2}: \{\textcolor{Blue}{R_{1}},\textcolor{Blue}{R_{2}}, \textcolor{ForestGreen}{R_{6}} \} \\
H_{3}: \{\textcolor{Blue}{R_{1}},\textcolor{Blue}{R_{2}},\textcolor{Blue}{R_{3}}, \textcolor{ForestGreen}{R_{6}} \} \\
\end{split}
\end{equation}

The fuzzy model with rule base $H_{true}$ consists of the rules $R_1-R_3$ and is the true model from which the data is generated. $H_1$ consists of two additional rules $R_4, R_5$ which are antagonistic to the original set of rules. In $H_2$ one of the original rules $R_3$ is replaced with $R_6$ which includes the dummy covariate. $H_3$ has all the rules as $H_{true}$ with the additional dummy rule $R_6$.

The 50 data points for comparing the evidence among the different models is generated by randomly sampling from $H_{true}$ with parameters $\phi_i=[5,5,5,5,5,5,50,50,50]$. The universe of discourse for the referential sets $loc\_risk,  maintenance,  dummy\_covar$ is $[0,10]$ and for $downtime$ is $[0,100]$. The $dummy\_covar$ is also randomly sampled within $[0,10]$ and appended to the input dataset.

We also compare the traditional GLMs with linear predictors to show their evidence on the same data-set. We use the Gaussian likelihood function as given in Eqn. (\ref{eq:normal:assumption}) for this. Each element of the mean vector for these GLMs are linear predictors of the form 

\begin{equation}
\begin{aligned}
\label{eq:GLM1}
GLM1: \mu=\alpha_0 + & \alpha_1 x_1 x_2 \\
GLM2: \mu=\alpha_0 + &\sum_{i=1}^2 \alpha_{i} x_{i} \\
GLM3: \mu=\alpha_0 +  &\sum_{i=1}^2 \alpha_{i} x_{i}^2 + \alpha_3 x_1 x_2 \\
GLM4: \mu=\alpha_0 +  &\sum_{i=1}^2 \alpha_{i} x_{i} +\sum_{i=1}^2 \alpha_{i+2} x_{i}^2 + \alpha_5 x_1 x_2 \\
GLM5: \mu= \alpha_0 + & \sum_{i=1}^2 \alpha_{i} x_{i} 
                                 + \sum_{i=1}^2 \alpha_{i+2} x_{i}^2 + \alpha_5 x_1 x_2\\ 
                                 +&\sum_{i=1}^2 \alpha_{i+5} x_{i}^3 + \alpha_8 x_1^2 x_2
\end{aligned}
\end{equation}
where $\alpha_i \forall i$ are the parameters of the models. The range of all $\alpha_i$ for the GLMs is between -50 to 50. Since the evidence calculation is also dependent on the support of the prior, changing this range would influence the final evidence values. We have shown such simulations later for the real world case.

The evidence calculations from these multiple models for the synthetic dataset is shown in Table \ref{table_comparsion} using two different implementations of the nested sampling algorithm - single and multi-ellipsoidal methods. The corresponding mean and standard deviation of the posterior distributions of the model parameters are shown in Table \ref{posterior_synthetic}. 

Both models $H_2$ and $H_3$ have three additional parameters which estimates the hyper-parameters of the membership function of the dummy variable. The posterior values of these are not shown in the appendix in Table \ref{posterior_synthetic} as they cannot be compared with respect to a true value. All the simulations in the paper are done using the python nested sampling library nestle \cite{nestle2016}. For both the synthetic and the real world case the number of particles or live points are set to 50 and the algorithm terminates when the difference between the estimated total log evidence and the current log evidence is less than the user defined threshold of 0.5.

\begin{table}[!th]
\renewcommand{\arraystretch}{1.3}
\caption{Comparison of model evidences for all the synthetic dataset case}
\label{table_comparsion}
\centering

\begin{tabular}{|c|c|c|c|c|}
\hline
& \multicolumn{2}{c|}{Single Ellipsoidal}  & \multicolumn{2}{c|}{Multi Ellipsoidal} \\

Model & $log(Z)$ & Fcalls & $log(Z)$ & Fcalls \\
\hline
$H_{true}$ & -72.087 $\pm$ 0.665 & 27387 & -71.880 $\pm$ 0.657 & 13257 \\
\hline
$H_{1}$ & -738.822 $\pm$ 0.714 & 9127 & -738.699 $\pm$ 0.710 & 7495 \\
\hline
$H_{2}$ & -460.352 $\pm$ 0.672 & 14862 & -460.362 $\pm$ 0.676 & 8586 \\
\hline
$H_{3}$ & -71.867 $\pm$ 0.658 & 11631 & -72.878 $\pm$ 0.677 & 11631 \\
\hline
$H_{GLM1}$ & -3564.282 $\pm$ 0.598 & 1904 & -3564.282 $\pm$ 0.598 & 1904 \\
\hline
$H_{GLM2}$ & -1330.995 $\pm$ 0.623 & 2510 & -1330.995 $\pm$ 0.623 & 2510 \\
\hline
$H_{GLM3}$ & -1547.996 $\pm$ 0.779 & 6978 & -1548.437 $\pm$ 0.782 & 5220 \\
\hline
$H_{GLM4}$ & -1269.334 $\pm$ 0.909 & 7529 & -1268.904 $\pm$ 0.903 & 10612 \\
\hline
$H_{GLM5}$ & -694.100 $\pm$ 1.130 & 26739 & -693.865 $\pm$ 1.128 & 63324 \\
\hline

\end{tabular}
\end{table}

Table \ref{table_comparsion} shows that the model $H_{true}$ has the highest value of evidence $(log(Z))$, with $H_3$ following close behind, using the multi-ellipsoidal method which is expected. However with the single ellipsoidal method, $H_3$ has a marginally lower $(log(Z))$ value than $H_{true}$ implying that $H_3$ is preferred. Now the error estimates in the $log(Z)$ indicates that this difference is within the error margin and indeed looking at Eqn. (\ref{eq:models}) we see that $H_3$ consists of all the rules of $H_{true}$ along with a rule composed of the dummy variable which would have negligible impact on the model output. Either the number of particles in the algorithm needs to be increased to discern this (this increases accuracy with a consequent increase in computational run time) or 50 data points might not be enough and more data is needed.  All the other fuzzy models ($H_1, H_3$) and GLMs have a much lower evidence value. Nevertheless, the evidence metric helps in quantitatively ranking the different alternative models and deciding how worse or better they are in comparison to each other. 
As also shown in Table \ref{table_comparsion}, the number of function calls (Fcalls) for the multi-ellipsoidal algorithm is generally lower or the same as that of single ellipsoidal algorithm except in some cases like $H_{GLM5}$. This implies that in cases where the likelihood function is not multi-modal and both the algorithms would give similar answers, the multi-ellipsoidal algorithm would be useful in terms of run time reduction, especially for expensive likelihood functions (e.g. fuzzy rule based models).

\subsection{A real world example}

We use the same dataset and problem setting of modelling insurance risks for power plants as described in \cite{pan2016fuzzy}. Please refer to \cite{pan2016fuzzy} for detailed background of the problem and the rationale for applying the Fuzzy Bayesian methodology. The key idea is to encode the underwriters opinions as a fuzzy rule base expressing dependence between the three covariates -- operation and maintenance (O\&M), loss history, design \& technology \& fire risk (DnTnF) which are used in predicting the system uptime and consequently helping in making insurance premium calculations and underwriting decisions where the dataset is very small. 

Table \ref{table_rule_base} shows three fuzzy models with different rule bases that are compared for the real world data-set using the evidence metric. $H_{rw1}$ is the same rule base as used in \cite{pan2016fuzzy} indicating that the underwriter perceives the O\&M to be more indicative of risk than the other covariates. $H_{rw2},H_{rw3}$ indicate that the loss history and the DnTnF are the key covariates respectively.

\begin{table}[!th]
\renewcommand{\arraystretch}{1.3}
\caption{Rule bases for the different Fuzzy Bayesian models fitted on the real world case}
\label{table_rule_base}
\centering
\begin{tabular}{|c|c|c|c||c|}
\hline
\multicolumn{1}{|c}{Model} & \multicolumn{3}{|c||}{Antecedents}  & \multicolumn{1}{c|}{Consequent} \\
\cline{2-5}
& O\&M & Loss history &  DnTnF  & System uptime \\
\hline
\multirow{5}{*}{$H_{rw1}$} & Good & - & - & High \\
& Good & - & Average & Medium \\
& Average & - & - & High \\
& Bad & Good & - & High \\
& Bad & Bad & Bad & Low \\
\hline

\multirow{5}{*}{$H_{rw2}$} & - & Good & - & High \\
& - & Good & Average & Medium \\
& - & Average & - & High \\
& Good & Bad & - & Low \\
& Bad & Bad & Bad & Low \\
\hline

\multirow{5}{*}{$H_{rw3}$} & - & - & Good & High \\
& Average & - & Good & Medium \\
& - & - & Average & Medium \\
& - & Good & Bad & High \\
& Bad & Bad & Bad & Low \\
\hline

\end{tabular}
\end{table}

In addition we also compare three GLM models, as given by Eqn.\ (\ref{eq:GLM678}), on the real world data-set. 
Three sets of simulations are conducted for the real world case, where $\Sigma=\sigma^2 I_N$ in Eqn.\ (\ref{eq:normal:assumption}) is estimated from the data itself and  $\sigma$ is set arbitrarily to constant values of 0.25 and 1.0. The corresponding simulation results are reported in Tables \ref{real_table_comparsion}, \ref{real_table_comparsion_sigma_25} and \ref{real_table_comparsion_sigma_1} respectively. The tables also contain an additional column for the prior ranges. The fuzzy model parameters have been defined to be within the range of $[0,10]$, however the GLM parameters can be any real number and hence two different prior ranges are tested as choice of prior range influences the evidence values. As can be observed from Table \ref{real_table_comparsion}, an uniform prior over a narrower range $[-10,10]$ has a bit higher value of evidence than over a wider range of $[-100,100]$. This might be due to the fact that we are distributing the same probability mass over a narrower range (having more confidence in them as we are ruling out the other values) as opposed to a wider one. However, this is not necessarily the case when $\sigma$ is artificially set to a lower value in Table \ref{real_table_comparsion_sigma_25} (e.g. $H_{GLM7},H_{GLM8}$).

\begin{table}[!th]
\renewcommand{\arraystretch}{1.3}
\caption{Comparison of model evidences for the real world example: $\sigma$ estimated from parameters}
\label{real_table_comparsion}
\centering
\setlength{\tabcolsep}{2pt}
\begin{tabular}{|c|c|c|c|c|c|}
\hline
& & \multicolumn{2}{c|}{Single Ellipsoidal}  & \multicolumn{2}{c|}{Multi Ellipsoidal} \\

Model & Prior & $log(Z)$ & Fcalls & $log(Z)$ & Fcalls \\
\hline
$H_{rw1}$ & [0,10] & -59.314 $\pm$ 0.350 & 1464 & -59.314 $\pm$ 0.350 & 1464 \\
\hline
$H_{rw2}$ & [0,10] & -62.412 $\pm$ 0.281 & 969 & -62.412 $\pm$ 0.281 & 969 \\
\hline
$H_{rw3}$ & [0,10] & -66.469 $\pm$ 0.223 & 504 & -66.469 $\pm$ 0.223 & 504 \\
\hline
\multirow{2}{*}{$H_{GLM6}$} & [-100,100] & -74.591 $\pm$ 0.642 & 3533 & -73.679 $\pm$ 0.630 & 4584 \\
& [-10,10] & -65.368 $\pm$ 0.489 & 1898 & -65.339 $\pm$ 0.486 & 1849 \\
\hline
\multirow{2}{*}{$H_{GLM7}$} & [-100,100] & -95.391 $\pm$ 0.922 & 10420 & -94.500 $\pm$ 0.907 & 18289 \\
& [-10,10] & -79.761 $\pm$ 0.731 & 4500 & -78.925 $\pm$ 0.719 & 5876 \\
\hline
\multirow{2}{*}{$H_{GLM8}$} & [-100,100] & -109.288 $\pm$ 1.048 & 53316 & -109.722 $\pm$ 1.053 & 39006 \\
& [-10,10] & -86.613 $\pm$ 0.804 & 19700 & -87.005 $\pm$ 0.807 & 18399 \\
\hline
\hline

\end{tabular}
\end{table}

Table \ref{real_table_comparsion} shows that $H_{rw1}$ has the highest model evidence among the competing alternatives. Arbitrarily setting the value of $\sigma=1.0$ as done in Table \ref{real_table_comparsion_sigma_1} still has $H_{rw1}$ as the best model but the ranking of the other models change. This effect is more predominant in Table \ref{real_table_comparsion_sigma_25} where $\sigma$ is artificially set to an even lower value of 0.25 and $H_{GLM8}$ has the lowest value of evidence among all the competing alternatives.

\begin{table}[!th]
\renewcommand{\arraystretch}{1.3}
\caption{Comparison of model evidences for the real world example: $\sigma=0.25$ }
\label{real_table_comparsion_sigma_25}
\centering
\setlength{\tabcolsep}{1pt}
\begin{tabular}{|c|c|c|c|c|c|}
\hline
& & \multicolumn{2}{c|}{Single Ellipsoidal}  & \multicolumn{2}{c|}{Multi Ellipsoidal} \\

Model & Prior & $log(Z)$ & Fcalls & $log(Z)$ & Fcalls \\
\hline
$H_{rw1}$ & [0,10] & -596.783 $\pm$ 0.673 & 12454 & -598.993 $\pm$ 0.691 & 6392 \\
\hline
$H_{rw2}$ & [0,10] &  -780.824 $\pm$ 0.765 & 28171 & -780.058 $\pm$ 0.745 & 16852 \\
\hline
$H_{rw3}$ & [0,10] &  -1903.106 $\pm$ 0.642 & 10605 & -1904.153 $\pm$ 0.661 & 5648 \\
\hline
\multirow{2}{*}{$H_{GLM6}$} & [-100,100] &  -795.069 $\pm$ 0.735 & 5564 & -795.367 $\pm$ 0.741 & 4115 \\
& [-10,10] &  -787.166 $\pm$ 0.620 & 2222 & -786.612 $\pm$ 0.612 & 2358 \\
\hline
\multirow{2}{*}{$H_{GLM7}$} & [-100,100] &  -643.082 $\pm$ 1.077 & 18943 & -643.200 $\pm$ 1.073 & 20175 \\
& [-10,10] &  -652.957 $\pm$ 0.918 & 8948 & -652.871 $\pm$ 0.919 & 9109 \\
\hline
\multirow{2}{*}{$H_{GLM8}$} & [-100,100] &  -589.990 $\pm$ 1.237 & 546259 & -589.281 $\pm$ 1.237 & 1066203 \\
& [-10,10] & -598.993 $\pm$ 1.060 & 21860 & -597.475 $\pm$ 1.050 & 65359 \\
\hline
\hline

\end{tabular}
\end{table}

\begin{table}[!th]
\renewcommand{\arraystretch}{1.3}
\caption{Comparison of model evidences for the real world example: $\sigma=1.0$ }
\label{real_table_comparsion_sigma_1}
\centering
\setlength{\tabcolsep}{2pt}
\begin{tabular}{|c|c|c|c|c|c|}
\hline
& & \multicolumn{2}{c|}{Single Ellipsoidal}  & \multicolumn{2}{c|}{Multi Ellipsoidal} \\

Model & Prior & $log(Z)$ & Fcalls & $log(Z)$ & Fcalls \\
\hline
$H_{rw1}$ & [0,10] & -73.250 $\pm$ 0.449 & 5451 & -73.575 $\pm$ 0.463 & 3295 \\
\hline
$H_{rw2}$ & [0,10] & -88.356 $\pm$ 0.500 & 13774 & -88.637 $\pm$ 0.527 & 8447 \\
\hline
$H_{rw3}$ & [0,10] & -150.194 $\pm$ 0.330 & 1443 & -150.636 $\pm$ 0.323 & 929 \\
\hline
\multirow{2}{*}{$H_{GLM6}$} & [-100,100] & -94.155 $\pm$ 0.659 & 3888 & -94.727 $\pm$ 0.666 & 2966 \\
& [-10,10] &  -86.649 $\pm$ 0.532 & 1697 & -85.996 $\pm$ 0.518 & 1596 \\
\hline
\multirow{2}{*}{$H_{GLM7}$} & [-100,100] &  -110.170 $\pm$ 0.972 & 9579 & -110.018 $\pm$ 0.970 & 11597 \\
& [-10,10] &  -95.198 $\pm$ 0.775 & 6355 & -94.820 $\pm$ 0.771 & 5959 \\
\hline
\multirow{2}{*}{$H_{GLM8}$} & [-100,100] &   -123.227 $\pm$ 1.130 & 50185 & -123.128 $\pm$ 1.118 & 57867 \\
& [-10,10] &  -102.192 $\pm$ 0.895 & 13508 & -100.428 $\pm$ 0.876 & 32887 \\
\hline
\hline

\end{tabular}
\end{table}

\section{Discussions and challenges}

The example cases showed the model comparison between different rule bases arrived at by different thought processes by multiple experts who do not concur. The method however is much more generic and can also be used to choose other hyper-parameters in the fuzzy model as well. For example, if there is reason to believe that trapezoidal membership functions might be a better representation of the fuzzy sets than the triangular ones used here, then a model with the same rule base but with different shapes of membership functions can be used and the evidence of the model vis-\`a-vis the other one can decide which type of membership function is best supported by the data.  

One advantage of the method was that it presents a unified way for model comparison among widely different modelling philosophies (like the FBL models and the GLMs). This would not have been possible with other approximate methods like AIC, DIC or BIC as the number of `effective' parameters is difficult to find out and have different implications for the different models. Also the use of nested sampling directly produces the evidence along with the posterior parameter estimates making the parameter estimation and model comparison in a single step as opposed to a two stage process.

To lend more credibility to the rankings of the different models based on their evidence, it would be better to perform some sort of sensitivity analysis by checking how the rankings change depending on modelling assumptions like choice of prior, likelihood etc. If the results are not very sensitive to these variations, then the results are more reliable.

\section{Conclusion}
This paper looked at comparing different rule bases in Fuzzy Bayesian Learning using marginal likelihood. Synthetic and real world case studies show how the methodology is able to quantitatively rank the alternative rule bases and also the pitfalls of setting model parameters (like variance) arbitrarily instead of learning from the small dataset. The methodology is generic and can compare other types of models (e.g. GLMs) as demonstrated in the case studies. The marginal likelihood automatically trades-off between model complexity and predictive accuracy and does not need special parameters for tweaking unlike other information theoretic criteria and hence is easy to apply to a large set of models. Future work can look at automatically generating better and interpretable rule bases with higher evidence values using machine intelligence.

\section*{Acknowledgment}

The authors would like to thank Sciemus Ltd. for sponsoring this work.

\ifCLASSOPTIONcaptionsoff
  \newpage
\fi



%

\bibliographystyle{IEEEtran}
\bibliography{mybiblio}

\begin{thebibliography}{10}
\providecommand{\url}[1]{#1}
\csname url@samestyle\endcsname
\providecommand{\newblock}{\relax}
\providecommand{\bibinfo}[2]{#2}
\providecommand{\BIBentrySTDinterwordspacing}{\spaceskip=0pt\relax}
\providecommand{\BIBentryALTinterwordstretchfactor}{4}
\providecommand{\BIBentryALTinterwordspacing}{\spaceskip=\fontdimen2\font plus
\BIBentryALTinterwordstretchfactor\fontdimen3\font minus
  \fontdimen4\font\relax}
\providecommand{\BIBforeignlanguage}[2]{{%
\expandafter\ifx\csname l@#1\endcsname\relax
\typeout{** WARNING: IEEEtran.bst: No hyphenation pattern has been}%
\typeout{** loaded for the language `#1'. Using the pattern for}%
\typeout{** the default language instead.}%
\else
\language=\csname l@#1\endcsname
\fi
#2}}
\providecommand{\BIBdecl}{\relax}
\BIBdecl

\bibitem{pan2016fuzzy}
I.~Pan and D.~Bester, ``Fuzzy bayesian learning,'' \emph{arXiv preprint
  arXiv:1610.09156}, 2016.

\bibitem{juang2014rule}
C.-F. Juang, C.-W. Hung, and C.-H. Hsu, ``Rule-based cooperative continuous ant
  colony optimization to improve the accuracy of fuzzy system design,''
  \emph{IEEE Transactions on Fuzzy Systems}, vol.~22, no.~4, pp. 723--735,
  2014.

\bibitem{cara2013multiobjective}
A.~B. Cara, C.~Wagner, H.~Hagras, H.~Pomares, and I.~Rojas, ``Multiobjective
  optimization and comparison of nonsingleton type-1 and singleton interval
  type-2 fuzzy logic systems,'' \emph{IEEE Transactions on Fuzzy systems},
  vol.~21, no.~3, pp. 459--476, 2013.

\bibitem{gil2015gain}
P.~Gil, C.~Lucena, A.~Cardoso, and L.~B. Palma, ``Gain tuning of fuzzy pid
  controllers for mimo systems: A performance-driven approach,'' \emph{IEEE
  Transactions on Fuzzy Systems}, vol.~23, no.~4, pp. 757--768, 2015.

\bibitem{zhao2013hybrid}
W.~Zhao, Q.~Niu, K.~Li, and G.~W. Irwin, ``A hybrid learning method for
  constructing compact rule-based fuzzy models,'' \emph{IEEE transactions on
  cybernetics}, vol.~43, no.~6, pp. 1807--1821, 2013.

\bibitem{othman2014efis}
A.~A. Othman, H.~R. Tizhoosh, and F.~Khalvati, ``Efis—evolving fuzzy image
  segmentation,'' \emph{IEEE Transactions on Fuzzy Systems}, vol.~22, no.~1,
  pp. 72--82, 2014.

\bibitem{lin2015interval}
C.-T. Lin, N.~R. Pal, S.-L. Wu, Y.-T. Liu, and Y.-Y. Lin, ``An interval type-2
  neural fuzzy system for online system identification and feature
  elimination,'' \emph{IEEE transactions on neural networks and learning
  systems}, vol.~26, no.~7, pp. 1442--1455, 2015.

\bibitem{zhou2008low}
S.-M. Zhou and J.~Q. Gan, ``Low-level interpretability and high-level
  interpretability: a unified view of data-driven interpretable fuzzy system
  modelling,'' \emph{Fuzzy Sets and Systems}, vol. 159, no.~23, pp. 3091--3131,
  2008.

\bibitem{piironen2015comparison}
J.~Piironen and A.~Vehtari, ``Comparison of bayesian predictive methods for
  model selection,'' \emph{Statistics and Computing}, pp. 1--25, 2015.

\bibitem{vehtari2012survey}
A.~Vehtari, J.~Ojanen \emph{et~al.}, ``A survey of bayesian predictive methods
  for model assessment, selection and comparison,'' \emph{Statistics Surveys},
  vol.~6, pp. 142--228, 2012.

\bibitem{lunn2012bugs}
D.~Lunn, C.~Jackson, N.~Best, A.~Thomas, and D.~Spiegelhalter, \emph{The BUGS
  book: A practical introduction to Bayesian analysis}.\hskip 1em plus 0.5em
  minus 0.4em\relax CRC press, 2012.

\bibitem{murphy2012machine}
K.~P. Murphy, \emph{Machine learning: a probabilistic perspective}.\hskip 1em
  plus 0.5em minus 0.4em\relax MIT press, 2012.

\bibitem{schoniger2014model}
A.~Sch{\"o}niger, T.~W{\"o}hling, L.~Samaniego, and W.~Nowak, ``Model selection
  on solid ground: Rigorous comparison of nine ways to evaluate bayesian model
  evidence,'' \emph{Water resources research}, vol.~50, no.~12, pp. 9484--9513,
  2014.

\bibitem{skilling2004nested}
J.~Skilling, ``Nested sampling,'' \emph{Bayesian inference and maximum entropy
  methods in science and engineering}, vol. 735, pp. 395--405, 2004.

\bibitem{mackay2003information}
D.~J. MacKay, \emph{Information theory, inference and learning
  algorithms}.\hskip 1em plus 0.5em minus 0.4em\relax Cambridge university
  press, 2003.

\bibitem{smith1980bayes}
A.~F. Smith and D.~J. Spiegelhalter, ``Bayes factors and choice criteria for
  linear models,'' \emph{Journal of the Royal Statistical Society. Series B
  (Methodological)}, pp. 213--220, 1980.

\bibitem{jefferys1992ockham}
W.~H. Jefferys and J.~O. Berger, ``Ockham's razor and bayesian analysis,''
  \emph{American Scientist}, vol.~80, no.~1, pp. 64--72, 1992.

\bibitem{mackay1992bayesian}
D.~J. MacKay, ``Bayesian interpolation,'' \emph{Neural computation}, vol.~4,
  no.~3, pp. 415--447, 1992.

\bibitem{feroz2008multimodal}
F.~Feroz and M.~Hobson, ``Multimodal nested sampling: an efficient and robust
  alternative to markov chain monte carlo methods for astronomical data
  analyses,'' \emph{Monthly Notices of the Royal Astronomical Society}, vol.
  384, no.~2, pp. 449--463, 2008.

\bibitem{mukherjee2006nested}
P.~Mukherjee, D.~Parkinson, and A.~R. Liddle, ``A nested sampling algorithm for
  cosmological model selection,'' \emph{The Astrophysical Journal Letters},
  vol. 638, no.~2, p. L51, 2006.

\bibitem{handley2015polychord}
W.~Handley, M.~Hobson, and A.~Lasenby, ``Polychord: nested sampling for
  cosmology,'' \emph{Monthly Notices of the Royal Astronomical Society:
  Letters}, vol. 450, no.~1, pp. L61--L65, 2015.

\bibitem{brewer2011diffusive}
B.~J. Brewer, L.~B. P{\'a}rtay, and G.~Cs{\'a}nyi, ``Diffusive nested
  sampling,'' \emph{Statistics and Computing}, vol.~21, no.~4, pp. 649--656,
  2011.

\bibitem{nestle2016}
``nestle, pure python, mit-licensed implementation of nested sampling
  algorithms,'' \url{http://kbarbary.github.io/nestle/}, accessed: 2016-12-22.

\end{thebibliography}
\vfill

\appendices

\section{}

\begin{table*}[!hb]
\renewcommand{\arraystretch}{1.3}
\scriptsize
\centering
\caption{Posterior parameter estimates for the synthetic simulation cases, mean (standard deviation)}
\label{posterior_synthetic}
\begin{tabular}{|c|c|c|c|c|c|c|c|c|c|c|c|c|}
\hline
Model & Method & theta0 & theta1 & theta2 & theta3 & theta4 & theta5 & theta6 & theta7 & theta8 \\
\hline
\multirow{ 2}{*}{$H_{true}$} & Single Ellip & 4.98 (0.12) &   4.99 (0.21) &   5.02 (0.07) &   5.02 (0.21) &   4.97 (0.21) &   4.95 (0.26) &  49.22 (4.46) &  50.06 (0.98) &  49.85 (2.11) \\
\cline{2-11}
& Multi Ellip &  4.98 (0.12) &   4.98 (0.22) &   5.03 (0.07) &   5.04 (0.23) &   4.95 (0.22) &   4.97 (0.25) &  49.14 (4.64) &  49.96 (1.00) &  49.74 (2.34) \\
\hline
\multirow{ 2}{*}{$H_{1}$} & Single Ellip &  0.48 (0.28) &  0.14 (0.13) &   3.66 (0.09) &   9.95 (0.04) &   9.13 (0.12) &   3.68 (2.07) &   3.16 (0.29) &  18.64 (1.16) &  72.28 (1.30) \\
\cline{2-11}
& Multi Ellip &  0.49 (0.29) &   0.13 (0.12) &   3.68 (0.09) &   9.95 (0.05) &   9.16 (0.15) &   3.79 (2.02) &   3.14 (0.29) &  18.71 (1.19) &  72.32 (1.42) \\

\hline

\multirow{ 2}{*}{$H_{2}$} & Single Ellip &  5.08 (2.81) &  3.92 (0.30) &   4.63 (0.11) &   7.50 (0.34) &   4.81 (0.18) &   4.98 (2.89) &   3.90 (0.27) &  43.26 (1.90) &  52.13 (1.96) \\
\cline{2-11}
& Multi Ellip & 4.95 (2.86) & 3.91 (0.27) & 4.63 (0.10) & 7.46 (0.37) & 4.80 (0.17) & 5.04 (2.78) & 3.91 (0.27) & 43.42 (2.01) &  51.94 (1.96) \\
\hline

\multirow{ 2}{*}{$H_{3}$} & Single Ellip &  4.98 (0.11) &  4.97 (0.23) &   5.03 (0.07) &   5.05 (0.22) &   4.95 (0.20) &   4.98 (0.26) &  49.21 (4.53) &  50.07 (1.05) &  49.78 (2.10) \\
\cline{2-11}
& Multi Ellip &  4.99 (0.12) &   4.98 (0.21) &   5.03 (0.07) &   5.04 (0.21) &   4.94 (0.19) &   4.96 (0.25) &  49.31 (4.42) &  50.08 (1.01) &  50.03 (2.10) \\
\hline

\multirow{ 2}{*}{$H_{GLM1}$} & Single Ellip & 49.99 (0.01) &   0.12 (0.00) &  - & - & - & - & - & - & - \\
\cline{2-11}
& Multi Ellip  & 49.99 (0.01) &   0.12 (0.00) &  - & - & - & - & - & - & - \\
\hline

\multirow{ 2}{*}{$H_{GLM2}$} & Single Ellip & 49.94 (0.06) &   2.83 (0.04) &  -1.91 (0.04) &  -  & -  & -  & -  & -  & - \\
\cline{2-11}
& Multi Ellip & 49.94 (0.06) &   2.83 (0.04) &  -1.91 (0.04) &  -  & -  & -  & -  & -  & - \\
\hline

\multirow{ 2}{*}{$H_{GLM3}$} & Single Ellip & 49.98 (0.02) &   0.24 (0.01) &  -0.18 (0.01) &   0.11 (0.02) &  - & - & - & - & - \\
\cline{2-11}
& Multi Ellip & 49.98 (0.02) &   0.24 (0.01) &  -0.18 (0.01) &   0.11 (0.02) &  - & - & - & - & - \\
\hline

\multirow{ 2}{*}{$H_{GLM4}$} & Single Ellip & 49.97 (0.03) &   3.69 (0.16) &  -3.62 (0.17) &  -0.11 (0.02) &   0.12 (0.02) &   0.13 (0.02) &  - & - & - \\
\cline{2-11}
& Multi Ellip & 49.97 (0.03) &   3.67 (0.16) &  -3.61 (0.16) &  -0.11 (0.02) &   0.12 (0.02) &   0.13 (0.02) &  - & - & - \\
\hline

\multirow{ 2}{*}{$H_{GLM5}$} & Single Ellip & 49.92 (0.08) &  16.70 (0.38) &  -10.86 (0.44) & -3.17 (0.10) &   1.73 (0.11) &  -0.50 (0.06) &   0.18 (0.01) &  -0.10 (0.01) &   0.09 (0.01) \\
\cline{2-11}
& Multi Ellip & 49.92 (0.08) &  16.64 (0.38) &  -10.82 (0.40) & -3.16 (0.10) &   1.71 (0.10) &  -0.49 (0.06) &   0.18 (0.01) &  -0.09 (0.01) &   0.09 (0.01) \\
\hline


\hline
\end{tabular}
\end{table*}

GLMs for the real-world dataset.
\begin{equation}
\begin{aligned}
\label{eq:GLM678}
\begin{split}
GLM6: \boldsymbol{\mu}=\alpha_0 + &\sum_{i=1}^3 \alpha_{i} x_{i} \\
GLM7: \boldsymbol{\mu}=\alpha_0 +&\sum_{i=1}^3 \alpha_{i} x_{i} + \sum_{\substack{
            i,j=1\\
            i \neq j}}^3 \alpha_{i+3} x_{i}x_{j} \\
            +& \sum_{\substack{
            i,j,k=1\\
            i \neq j \neq k}}^3 \alpha_{i+6} x_{i}x_{j}x_{k} \\           
GLM8: \boldsymbol{\mu}=\alpha_0 +&\sum_{i=1}^3 \alpha_{i} x_{i} + \sum_{i,j=1}^3 \alpha_{i+3} x_{i}x_{j} \\
            +& \sum_{\substack{
            i,j,k=1\\
            i \neq j \neq k}}^3 \alpha_{i+9} x_{i}x_{j}x_{k} 
\end{split}
\end{aligned}
\end{equation}

%








\end{document}